\documentclass[10pt,twocolumn,letterpaper]{article}

\usepackage{iccv}
\usepackage{times}
\usepackage{epsfig}
\usepackage{graphicx}
\usepackage{amsmath}
\usepackage{amssymb}
\usepackage{multirow}
\usepackage{booktabs}
\usepackage{bbding}
\usepackage[vlined,boxed,commentsnumbered,linesnumbered,ruled]{algorithm2e}
\usepackage{caption}
\usepackage{subcaption}
\usepackage{diagbox}

\usepackage[pagebackref=false,breaklinks=true,letterpaper=true,colorlinks,urlcolor=blue,citecolor=blue,linkcolor=blue,bookmarks=false]{hyperref}

\iccvfinalcopy 

\def\iccvPaperID{7672} 

\ificcvfinal\fi

\begin{document}

\title{Learning to Track Objects from Unlabeled Videos}

\author{Jilai Zheng$^{1}$ \quad Chao Ma$^{1}$\thanks{~Corresponding author.} \quad Houwen Peng$^{2}$ \quad Xiaokang Yang$^{1}$\\
${}^{1}$ MoE Key Lab of Artificial Intelligence, AI Institute, Shanghai Jiao Tong University \\
${}^{2}$ Microsoft Research\\
{\tt\small \{zhengjilai,chaoma,xkyang\}@sjtu.edu.cn, houwen.peng@microsoft.com}
}

\maketitle
\ificcvfinal\thispagestyle{empty}\fi


\begin{abstract}

In this paper, we propose to learn an Unsupervised Single Object Tracker (USOT) from scratch. We identify that three major challenges, i.e., moving object discovery, rich temporal variation exploitation, and online update, are the central causes of the performance bottleneck of existing unsupervised trackers. To narrow the gap between unsupervised trackers and supervised counterparts, we propose an effective unsupervised learning approach composed of three stages. First, we sample sequentially moving objects with unsupervised optical flow and dynamic programming, instead of random cropping. Second, we train a naive Siamese tracker from scratch using single-frame pairs. Third, we continue training the tracker with a novel cycle memory learning scheme, which is conducted in longer temporal spans and also enables our tracker to update online. Extensive experiments show that the proposed USOT learned from unlabeled videos performs well over the state-of-the-art unsupervised trackers by large margins, and on par with recent supervised deep trackers. Code is available at \url{https://github.com/VISION-SJTU/USOT}.

\vspace{-0.2in}

\end{abstract}

\section{Introduction}\label{sec:introduction}

Visual object tracking is one of the most fundamental computer vision tasks with numerous applications, such as autonomous driving, intelligent surveillance, robot, and human-computer interaction. The past few years have witnessed considerable progress in visual object tracking, thanks to the powerful representation of deep learning. In spite of the success, the state-of-the-art deep tracking algorithms are data-hungry, requiring a huge number of annotated data for supervised training. As manually labeled data are expensive and time-consuming, exploiting ubiquitous unlabeled videos for visual tracking has drawn increasing attention recently. 
Following the classic pipeline of unsupervised learning, existing unsupervised trackers randomly crop template regions on unlabeled videos and employ either self consistency~\cite{DBLP:conf/mm/SioMSCC20} or cycle consistency~\cite{DBLP:conf/cvpr/WangS0ZLL19} as a pretext task for learning to track. Despite the promising results, there still exists a large performance gap between unsupervised and supervised trackers.
In view of the great success of unsupervised learning on a number of other vision tasks, such as video object segmentation~\cite{DBLP:conf/cvpr/LaiLX20}, optical flow~\cite{DBLP:conf/cvpr/LiuZHLWTLWLH20} and depth estimation~\cite{DBLP:conf/cvpr/GodardAB17}, it is of great interest to narrow the gap between unsupervised and supervised trackers.

  \begin{figure}
	\begin{center}
		\includegraphics[width=0.48\textwidth]{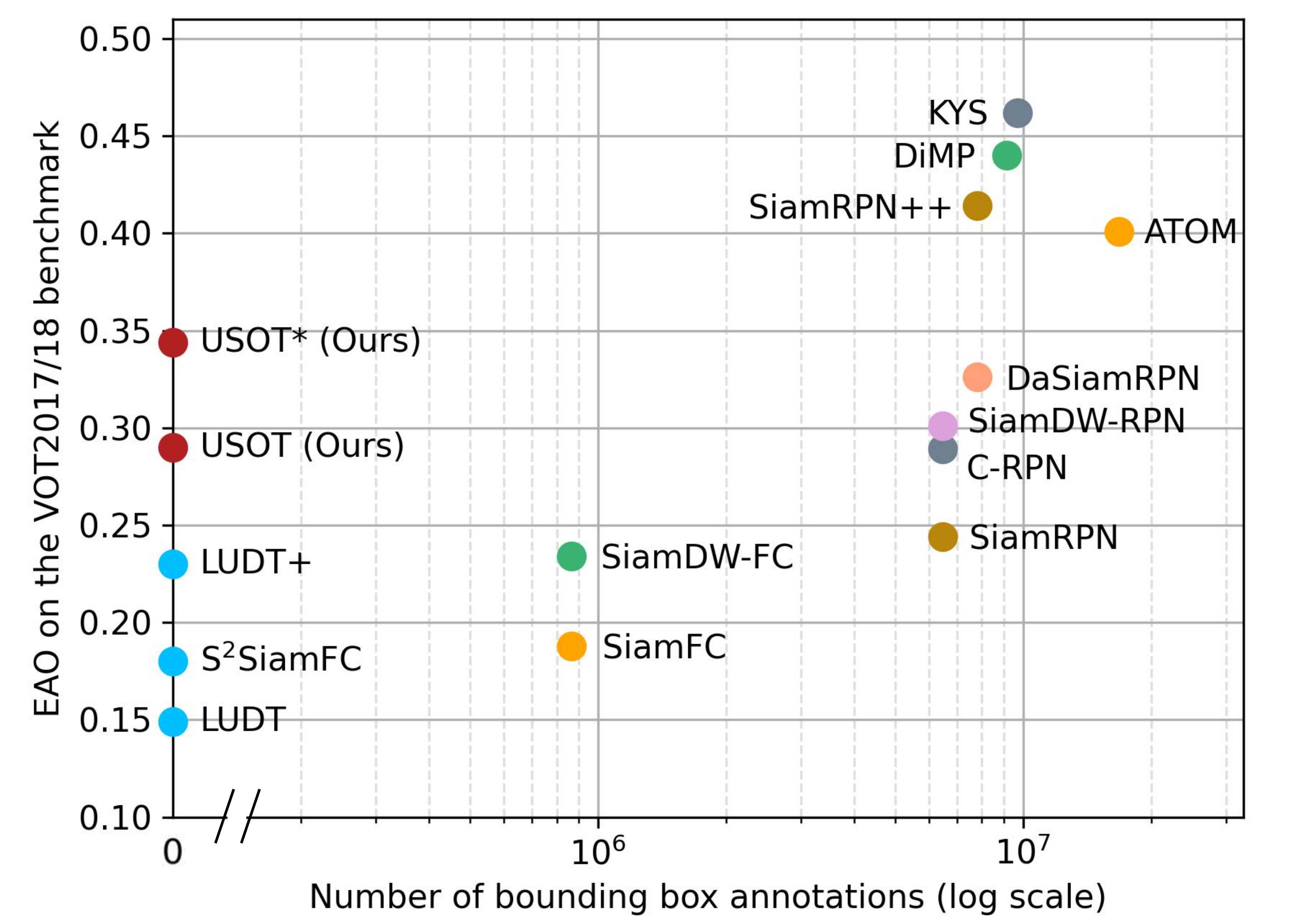}
	\end{center}
	\vspace{-0.25in}
	\caption{Comparison on the VOT2017/18 benchmark with recent deep trackers. The proposed trackers, USOT and USOT*, perform well over the state-of-the-art unsupervised deep trackers, including 	LUDT~\cite{DBLP:journals/ijcv/WangZSMLL21}, LUDT+~\cite{DBLP:journals/ijcv/WangZSMLL21}, and
	S$^2$SiamFC~\cite{DBLP:conf/mm/SioMSCC20}, and on par with recent supervised trackers. Notation:
	SiamFC~\cite{DBLP:conf/eccv/BertinettoVHVT16},
	SiamDW~\cite{DBLP:conf/cvpr/ZhangP19}, 
	SiamRPN~\cite{DBLP:conf/cvpr/LiYWZH18}, 
	C-RPN~\cite{DBLP:conf/cvpr/FanL19},
 	DaSiamRPN~\cite{DBLP:conf/eccv/ZhuWLWYH18}, 
	ATOM~\cite{DBLP:conf/cvpr/DanelljanBKF19},
	SiamRPN++~\cite{DBLP:conf/cvpr/LiWWZXY19},
	DiMP~\cite{DBLP:conf/iccv/BhatDGT19},
	KYS~\cite{DBLP:conf/eccv/BhatDGT20}.
	}
	\vspace{-0.2in}
	\label{fig:cover}
	
\end{figure}
 
We identify three critical challenges that cause the performance bottleneck of unsupervised trackers.
 \textit{1) Moving object discovery.} As ground truth bounding boxes are not available, existing unsupervised trackers randomly sample regions in frames as pseudo templates~\cite{DBLP:conf/cvpr/WangS0ZLL19, DBLP:conf/mm/SioMSCC20}. Random samples are far from precisely locating objects, not to mention learning to distinguish between objects and background. Moreover, as random samples do not contain clear edges of objects, they are not suitable for bounding box regression learning. The lack of bounding box regression for scale change estimation heavily limits the performance of unsupervised trackers. 	
 \textit{2) Rich temporal variation exploitation.} Due to the lack of labels in the temporal span, existing unsupervised trackers struggle to learn from rich motion clues. For example, UDT~\cite{DBLP:conf/cvpr/WangS0ZLL19} performs forward and backward tracking within at most $10$ frames. In such a short clip, the foreground objects show highly correlated appearances with little variations, causing a failure to exploit rich temporal variations over a long span for training. 
 \textit{3) Online update.} Online update helps to exploit the temporal smoothness and has demonstrated great success in leading supervised tracking methods~\cite{DBLP:conf/cvpr/Song0WGBZSL018, DBLP:conf/iccv/BhatDGT19, DBLP:conf/iccv/ZhangGWDK19, DBLP:conf/eccv/ZhangPFLH20}. 
 While supervised trackers usually collect multiple object samples in separated frames for learning online modules~\cite{DBLP:conf/iccv/BhatDGT19, STMTrack}, it is more challenging to train online branches for unsupervised trackers, due to lack of even coarse object locations in videos.

 To address these challenges, we propose to train a robust tracker from unlabeled videos. First, for data preparation, we develop a sequential box sampling algorithm to coarsely discover moving objects from unlabeled videos. Specifically, we use unsupervised optical flow to detect moving objects and apply dynamic programming to sequentially link candidate boxes. 
 Second, we naively train from scratch an unsupervised Siamese tracker using single-frame pairs. 
 That is, we train with each Siamese pair cropped based on the sampled box in a single frame.
 Despite its simplicity, we show that this strategy provides a great initialization for the unsupervised tracker, thus beneficial to future training in longer temporal spans.
 Third, we propose a cycle memory learning scheme to continue training the naive tracker. Specifically, we divide the whole video into a number of fragments according to the detected moving object trajectory. We then conduct forward tracking from a single frame to several other frames in the same fragment, and store all intermediate tracking results in a memory queue. We then track backward to the initial frame to compute the consistency loss. Since the length of video fragments is quite long (averaged $64.6$ frames on VID~\cite{DBLP:journals/ijcv/RussakovskyDSKS15}) compared with UDT~\cite{DBLP:conf/cvpr/WangS0ZLL19} ($<$10 frames), our tracker can capture large motion and appearance variations. More importantly, the proposed cycle memory scheme allows updating the memory queue online for model update (see Sec.~\ref{subsec:detail}). 
 
 We evaluate the proposed unsupervised tracker on six large-scale benchmarks. Extensive experiments show that our proposed tracker performs well over the state-of-the-art unsupervised trackers by large margins, and on par with the recent supervised trackers (see Fig.~\ref{fig:cover}). The main contributions of this work are summarized as follows:
 \vspace{-1.5mm}
 \begin{itemize}
 \setlength{\itemsep}{0em}
     \item We coarsely discover moving objects from unlabeled videos for unsupervised learning. 
     \item We train a naive Siamese tracker with single-frame pairs and gradually extend it to longer temporal spans.
     \item We propose a cycle memory learning scheme, allowing unsupervised trackers to update online. 
 \end{itemize}
 
\section{Related Work}

{\flushleft\bf Supervised Tracking.} Deep learning has revolutionized the field of visual tracking by powerful representation. In the past few years, template-based deep trackers with Siamese networks have received increasing attention due to the promising results on benchmark datasets. These trackers regard the target object as a template and search over a cropped window to locate the target. 
SiamFC~\cite{DBLP:conf/eccv/BertinettoVHVT16} first utilizes the same backbone network to extract deep features from both the template patch and the search patch, and computes the response map with cross-correlation. Since then, considerable efforts have been made to extend Siamese trackers. SiamRPN~\cite{DBLP:conf/cvpr/LiYWZH18} incorporates a region proposal network (RPN)~\cite{DBLP:conf/nips/RenHGS15} into the Siamese framework. DiMP~\cite{DBLP:conf/iccv/BhatDGT19} proposes to attach an online module to Siamese trackers for template update. Other noticeable improvements in the Siamese framework involve advanced backbone network~\cite{DBLP:conf/cvpr/ZhangP19}, correlation method~\cite{DBLP:conf/cvpr/LiWWZXY19}, attention mechanism~\cite{DBLP:conf/cvpr/YuXHS20}, re-detection module~\cite{DBLP:conf/cvpr/VoigtlaenderLTL20}, mask generation~\cite{DBLP:conf/cvpr/Wang0BHT19}, feature alignment~\cite{DBLP:conf/eccv/ZhangPFLH20}, anchor-free regressor~\cite{DBLP:conf/cvpr/ChenZLZJ20, DBLP:conf/cvpr/GuoWC0C20}, etc. With these efforts, Siamese trackers have shown the superior tracking performance thus far. 
However, training Siamese trackers requires a huge number of labeled training data. In this work, we aim at a novel unsupervised learning scheme, which helps to learn template-based trackers from unlabeled videos in the wild.  

\vspace{-3mm}
{\flushleft\bf Unsupervised Tracking.} The pioneer unsupervised deep tracker (UDT)~\cite{DBLP:conf/cvpr/WangS0ZLL19} suggests that a robust tracker is able to track an object forward and backward in a video and finally return to its initial location in the starting frame. UDT develops a tracker based on DCFNet~\cite{DBLP:journals/corr/WangGXZH17}, and computes a cycle consistency loss between forward and backward trajectories in the training phase. The contemporary method of UDT is TimeCycle~\cite{DBLP:conf/cvpr/WangJE19}, which proposes cycle consistency to generate unsupervised video representation. JSLTC~\cite{DBLP:conf/nips/LiLMWK019} proposes to calculate an inter-frame affinity matrix to model the transitions between video frames and use such correspondence to track objects. S$^2$SiamFC~\cite{DBLP:conf/mm/SioMSCC20} adopts the self-Siamese pipeline to train a foreground/background classifier like SiamFC~\cite{DBLP:conf/eccv/BertinettoVHVT16} from single-frame pairs, showing comparable results to the supervised counterpart. 
Despite the promising results, there exists a large performance gap between the state-of-the-art unsupervised trackers and the top-performing supervised ones. We identify three critical challenges, moving object discovery, rich temporal variation exploitation, and online update, that cause the performance bottleneck of unsupervised trackers. By effectively tackling these challenges, the proposed unsupervised tracker outperforms the state-of-the-art unsupervised trackers by large margins, and is on par with recent supervised trackers. 

\section{Proposed Method}

In this section, we present the proposed unsupervised tracker in detail. The unsupervised training scheme involves three stages. The first stage in Sec.~\ref{subsec:ufdp} aims to produce a trajectory of moving objects from unlabeled videos. The second stage in Sec.~\ref{subsec:selfsiamese} learns a naive Siamese tracker using single-frame pairs. The third stage in Sec.~\ref{subsec:cyclememory} continues training the naive tracker by means of cycle memory learning, which is performed in longer temporal spans and also enables the unsupervised tracker to update online. 

\subsection{Moving Object Discovery}\label{subsec:ufdp}

Instead of randomly cropping objects, we propose to generate a smooth bounding box sequence for moving foreground objects in unlabeled videos. For discovering moving objects, we have two key observations:
\begin{itemize}
\setlength{\itemsep}{0em}
    \item Foreground objects tend to have distinguishing motion patterns in contrast to the background surroundings. This inspires us to discover candidate foreground objects by means of unsupervised optical flow.
    \item The trajectories of moving objects tend to be smooth. This motivates us to employ dynamic programming (DP) to get temporally reliable box sequences.
\end{itemize}

\begin{figure}
	\begin{center}
		\includegraphics[width=0.45\textwidth]{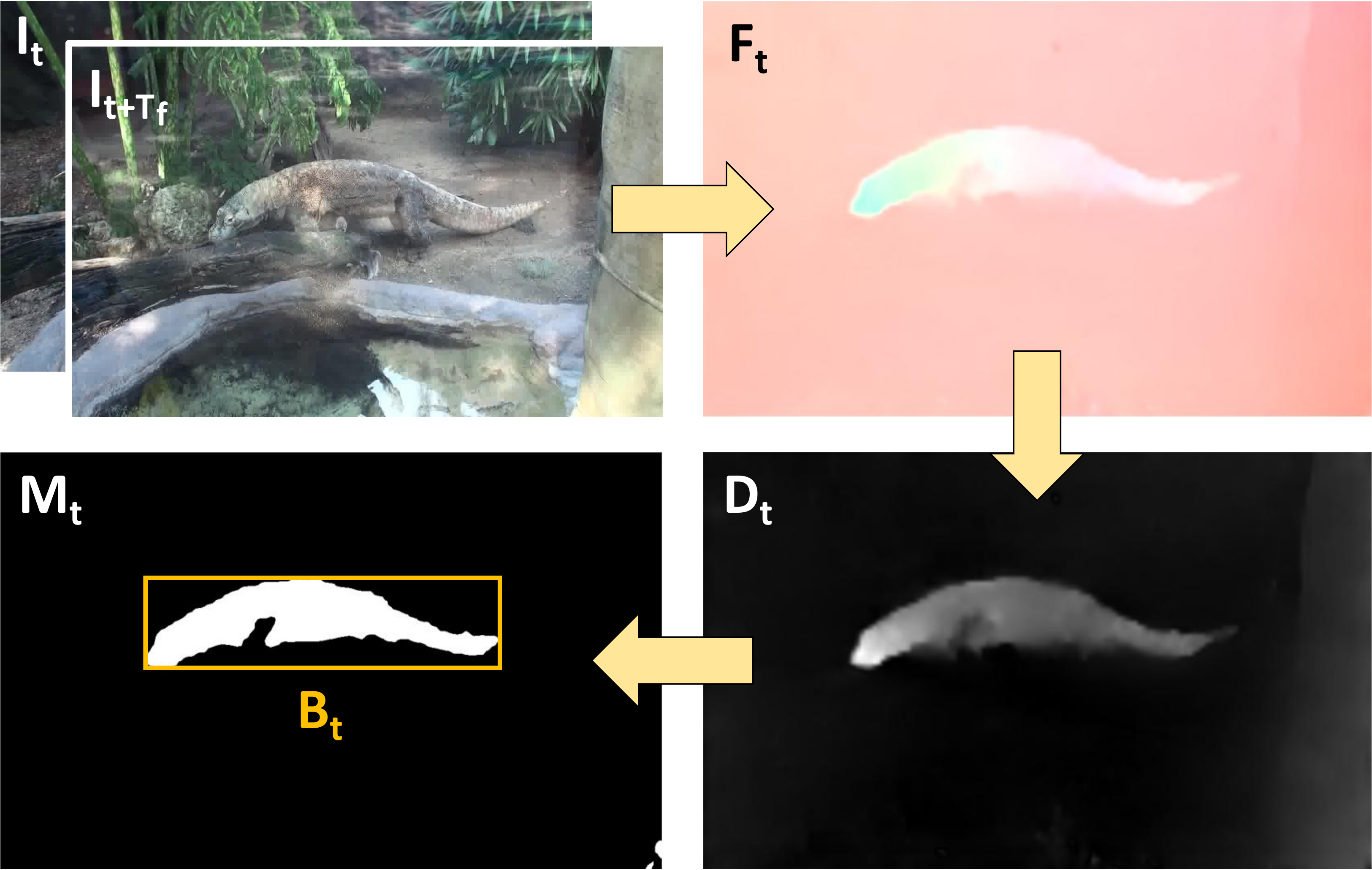}
	\end{center}
	\vspace{-0.15in}
	\caption{Candidate box generation via optical flow. The flow map $F_{t}$ contains the distinguishing motion patterns of moving objects. We binarize the flow map $F_{t}$ using a distance metric $D_t$ to generate the candidate box $B_t$.}
	\vspace{-0.15in}
	\label{fig:candidate_box}
\end{figure}

\vspace{-4mm}
{\flushleft\bf Candidate Box Generation.}
Let an arbitrary video be $\mathcal{I}$ including $L$ successive frames with the same size $W \times H$, namely $\mathcal{I} = \left\{ I_{t} \mid 1\le t \le L \right\}$, where $I_{t}$ is the $t^{th}$ frame in $\mathcal{I}$. To locate the potential foreground object in frame $I_{t}$, we first compute the optical flow map $F_{t}$ with frame $I_{t}$ and frame $I_{t+T_f}$, i.e., $F_{t} = {Flow}_{t \rightarrow t+T_f}$. $T_f$ is an interval for computing optical flow. 
As is illustrated in Fig.~\ref{fig:candidate_box}, we obtain $F_{t}$ from frame $I_{t}$ and frame $I_{t+T_f}$ using the off-the-shelf unsupervised ARFlow~\cite{DBLP:conf/cvpr/LiuZHLWTLWLH20} algorithm, and then transform $F_{t}$ to a distance map $D_{t}$. We binarize $D_{t}$ to get a mask $M_{t}$ as follows:  
\begin{equation}\label{eqn:binary}
\begin{aligned}
M_{t}^{i} = \left\{ \begin{array}{lc} 1 & \text{if } D_{t}^{i} \ge \alpha \cdot \max_{j}(D_{t}^{j}) + (1-\alpha) \cdot \operatorname{mean}_{j}(D_{t}^{j}) \\ 0 & \text{o.w.} \end{array} \right. \\ 
\text{where \ } D_{t}^{i} = \left\| F_{t}^{i} - \operatorname{mean}_{j}(F_{t}^{j})\right\|_{2}, \qquad \qquad \qquad &
\end{aligned} 
\end{equation}
and $\alpha\in(0,1)$ is a hyper-parameter. Here superscript denotes pixel-wise index. The maximum value and mean value within the spatial dimension are respectively indicated by $\max$ and $\operatorname{mean}$.

Every connected area with all internal pixels satisfying $M_{t}^{i} = 1$ corresponds to an area which has distinguishing motion compared with background in $I_{t}$, and this area is more likely to cover a foreground object. To further filter out unreliable areas from these candidates, we take the circumscribed rectangles of all these areas as initial candidate boxes, and score the boxes according to their sizes and positions. Due to center bias, larger bounding boxes in the middle of the image should have higher quality scores. Let $B = (x_0, y_0, x_1, y_1)$ denote the top-left and bottom-right corners of a box. The quality score $S_c$ of the box $B$ is defined as:
\begin{equation}\label{eqn:score}
\begin{aligned}
S_{c}(B) =  & (x_1-x_0)(y_1-y_0) + \\
& \beta \cdot \min(x_0, W - x_1)\min(y_0, H-y_1),
\end{aligned}
\end{equation}
where $\beta$ is a weight parameter. The box with the highest score is selected as the final candidate box $B_{t}$ for frame $I_{t}$. We denote the set of all these selected candidate boxes in video $\mathcal{I}$ as $\mathcal{B}=\left\{ B_{t} \mid 1 \le t \le L \right\}$.

\begin{figure*}
	\begin{center}
		\includegraphics[width=1.0\textwidth]{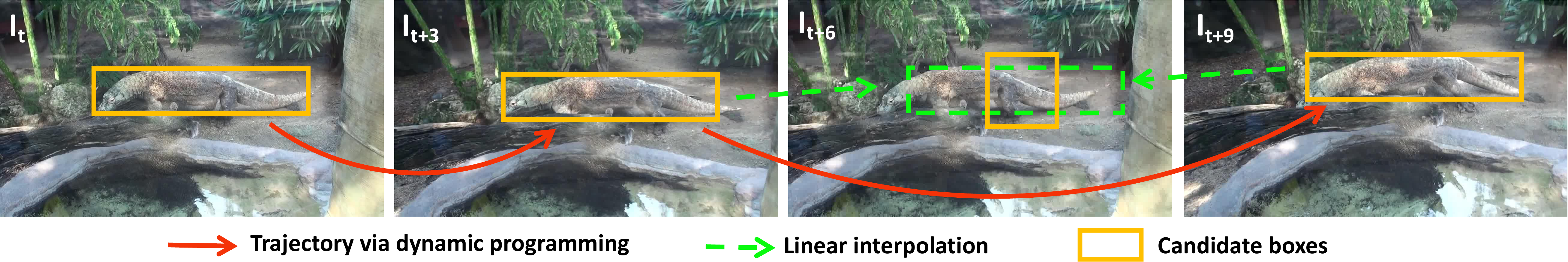}
	\end{center}
	\vspace{-0.22in}
	\caption{Box sequence generation. We use dynamic programming to generate a smooth and reliable box trajectory from candidate boxes in yellow. Pseudo boxes in green in the remaining frames are generated through linear interpolation.\label{fig:video_box} }
	\vspace{-0.10in}
\end{figure*}

\vspace{-2mm}
{\flushleft\bf Box Sequence Generation.}
The generated candidate bounding boxes $\mathcal{B}$ may contain noisy boxes due to camera shake, occlusion, etc. To remove unreliable boxes, we apply dynamic programming to create a more reliable bounding box sequence $\mathcal{B}^{\prime}$. According to the second observation that the trajectory of a moving object in a video should be smooth, we select a subset of candidate bounding boxes from $\mathcal{B}$, where the trajectory of the selected boxes is as smooth as possible. For dynamic programming, the most critical issue is how to measure the reward of transition in the box trajectory from one bounding box to another. We modify the DIoU~\cite{DBLP:conf/aaai/ZhengWLLYR20} metric, which originally considers the overlap and distance between two boxes. Formally, the reward $R_{dp}$ for dynamic programming is defined as:
\begin{equation}\label{eqn:reward}
	R_{dp}(B_{t}, B_{t^{\prime}}) = IoU(B_{t}, B_{t^{\prime}}) - \gamma \cdot R_{DIoU}(B_{t}, B_{t^{\prime}}),
\end{equation}
where $\gamma$ is a hyper-parameter. To encourage a smooth trajectory, we set $\gamma>1$ for the distance penalty on $R_{DIoU}$~\cite{DBLP:conf/aaai/ZhengWLLYR20}.  
Note that dynamic programming aims at discovering an optimal path in the box sequence $\mathcal{B}$ with the highest reward accumulation (see the supplementary document for the complete algorithm). 
As shown in Fig.~\ref{fig:video_box}, for the frames whose candidate boxes are not selected by DP, we use linear interpolation to generate pseudo boxes based on their adjacent candidate boxes selected by DP.

\subsection{Naive Siamese Tracker}\label{subsec:selfsiamese}

With the generated box sequences, we train a naive Siamese tracker using single-frame pairs from scratch. This pretext task is based on a simple observation that an image and any of its sub-region naturally form a training pair of the Siamese network~\cite{DBLP:conf/mm/SioMSCC20}. However, randomly sampled pseudo boxes as in \cite{DBLP:conf/mm/SioMSCC20} fail to cover foreground objects for effectively training Siamese networks. Moreover, random samples are not suitable for learning bounding box regression. This significantly hinders the performance of unsupervised trackers.
We propose to utilize the reliable box sequence $\mathcal{B}^{\prime}$ as training data. To ensure that the most precise bounding boxes in $\mathcal{B}^{\prime}$ are sampled by the data loader, we adopt a two-level scoring mechanism to filter out low-quality boxes at both the sequence and frame levels. We find that denser frame selection by DP in video $\mathcal{I}$ tends to imply a more successful moving object discovery. As such, we define the quality score $Q_{v}$ of video $\mathcal{I}=\left\{ I_{t} \mid 1 \le t \le L \right\}$ as: 
\begin{equation}\label{eqn:video_score}
	Q_{v}(\mathcal{I}) = \frac{N_{dp}}{L}, 
\end{equation}
where $N_{dp}$ indicates the number of frames in video $\mathcal{I}$ selected by DP. 

Similarly, the frame quality score evaluating the box $B^{\prime}_{t}$ in frame $I_{t}$ can be measured by the percentage of frames selected by DP within all its adjacent frames. Let $T_s$ be a fixed frame interval. We formally define the frame quality score $Q_{f}$ as: 
\begin{equation}\label{eqn:frame_score}
	Q_{f}(B^{\prime}_{t}) = \frac{N'_{dp}}{2T_s + 1},
\end{equation}
where $N'_{dp}$ indicates the number of frames between frame $I_{t-T_s}$ and frame $I_{t+T_s}$ selected by DP. 

When loading training pairs, we sequentially conduct video sampling and frame sampling. We sample a video only if its quality score $Q_v(\mathcal{I}) \ge \theta_{1}$, where $\theta_{1}$ is a threshold. During frame sampling, we randomly sample several frames with their total number positively correlated with $1/Q_{v}(\mathcal{I})$ from the selected video, and then select the frame with the highest frame quality score $Q_{f}(B^{\prime}_{t})$ for training. 

We follow the conventional training paradigm as in SiamFC~\cite{DBLP:conf/eccv/BertinettoVHVT16}. The input template $z_{t}$ and the search area $x_{t}$ are respectively of size $127 \times 127$ and $255 \times 255$, both cropped from $I_{t}$ based on $B^{\prime}_{t}$.
After extracting deep features from the input pair, we adopt PrPool~\cite{DBLP:conf/eccv/JiangLMXJ18} to pool the template feature, 
and then compute the multi-scale correlation map~\cite{DBLP:conf/eccv/ZhangPFLH20}. 
The output response map $\mathcal{R}_{cls}$ is of size $25 \times 25 \times 1$ for foreground/background classification. The other output response map $\mathcal{R}_{reg}$ is of size $25 \times 25 \times 4$ for regressing the distances from the center location to the four sides of the bounding box. The loss function $\mathcal{L}_{naive}$ is the sum of both the regression and classification losses: 
\begin{equation}\label{eqn:loss_stage2}
	\mathcal{L}_{naive} =  \mathcal{L}_{reg} + \lambda_{1}\mathcal{L}_{cls},
\end{equation}
where $\mathcal{L}_{reg}$ and $\mathcal{L}_{cls}$ are respectively the IoU loss~\cite{DBLP:conf/mm/YuJWCH16} and the binary cross-entropy (BCE) loss~\cite{DBLP:journals/anor/BoerKMR05}. $\lambda_{1}$ is a weight parameter.

\subsection{Cycle Memory Training}\label{subsec:cyclememory}

\begin{figure*}
	\begin{center}
		\includegraphics[width=1.0\textwidth]{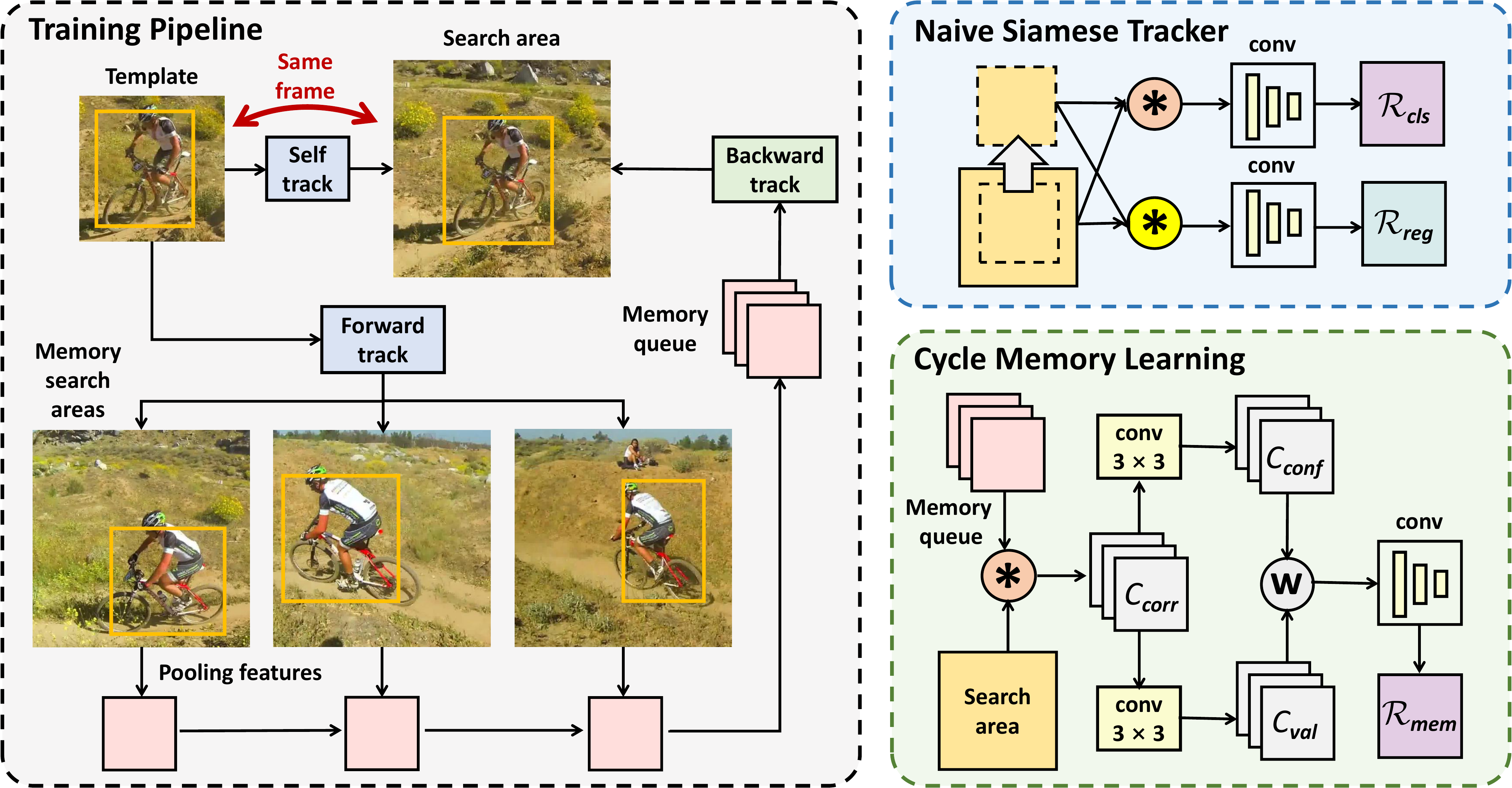}
	\end{center}
	\vspace{-0.15in}
	\caption{Overview of the proposed unsupervised tracking framework. Left: The overall training pipeline. Right: The detailed structure of the naive Siamese tracker for \textit{self tracking} and \textit{forward tracking}, and the online module learned with the cycle memory scheme. The naive tracker is trained with a template and a search area cropped from the same frame, while the online module aims to \textit{track backward} from the memory search areas to the template frame following the cycle learning pipeline. The circle notations with $*$ denote multi-scale correlation \cite{DBLP:conf/eccv/ZhangPFLH20} for deep features, where the same color indicates weight sharing. The circle with $W$ refers to the confidence-value module for integrating the correlation maps (Eqn.~\ref{eqn:integration}). 
	}

	\vspace{-0.15in}
	\label{fig:network}
\end{figure*}

We view the above unsupervised Siamese tracker as a naive tracker as it incurs two limitations. First, as the template and search area are cropped in the same frame, the tracker is not learned with large motion and appearance variations. Second, this tracker cannot update itself online, thus fails to track objects in long temporal spans or under complex scenes. 

We propose to continue training the naive tracker using a cycle memory learning scheme, aiming to enable the tracker to handle large variations as well as update the memory queue online. The main idea of cycle memory can be summarized as in Fig.~\ref{fig:network}. In brief, we first conduct forward tracking from a template $z_{t}$ to $N_{mem}$ adjacent memory frames, then store features of all intermediate tracking results as a memory queue, and finally conduct backward tracking to the original search area $x_t$. A cycle memory loss $\mathcal{L}_{mem}$ is computed using the same ground truth as $\mathcal{L}_{cls}$. 

Specifically, at every training step, we simultaneously crop a training pair $z_t$ and $x_t$ in frame $I_t$ (the same as training the naive Siamese tracker), as well as $N_{mem}$ memory search areas sampled from $\left\{x_{t} \mid T_l \le t \le T_u \right\}$. These memory search areas are cropped from $N_{mem}$ adjacent frames of $I_t$ according to the box sequence $\left\{B^{\prime}_{t} \mid T_l \le t \le T_u \right\}$. Here $T_l$ and $T_u$ are the lower and upper frame indices for sampling memory frames. Selecting these two indices is quite important. To learn from long-term variations, the frame interval between $T_l$ and $T_u$ should be large enough. However, excessive frame interval does harm to the learning process as the target object may disappear in frames far from $I_t$. In practice, we 
dynamically set $T_l$ and $T_u$ at frame $I_t$. Since they are two mirror variables, we formally define $T_u$ as follows:
\begin{equation}\label{eqn:upperbound}
\begin{aligned}
   T_u(I_t) &= \max_{t \le k \le L} \left\{ k \right\} \\ 
   \text{ s.t.\quad} &  \forall t < t^{\prime} \le k, R_{dp}(B^{\prime}_{t^{\prime}-1}, B^{\prime}_{t^{\prime}}) \ge \theta_{2} \\
   & \forall t < t^{\prime} \le k, Q_{f}(B^{\prime}_{t^{\prime}}) \ge \theta_{3},
\end{aligned}
\end{equation}
where $\theta_{2}$ and $\theta_{3}$ are two thresholds. The main idea of Eqn.~\ref{eqn:upperbound} is that, as long as a box $B^{\prime}_{k}$ can be connected to $B^{\prime}_{t}$ through a smooth and reliable box sequence in $\mathcal{B}^{\prime}$, search area cropped from $B^{\prime}_{k}$ can be used for cycle memory training. In other words, we take step changes in $\mathcal{B}^{\prime}$ to divide $\mathcal{I}$ into fragments, and the pseudo boxes of all frames in the same fragment tend to locate the same object. This scheme helps our tracker exploit long-term variations, while still ensuring the reliability of pseudo bounding boxes in the memory frames (see Sec.~\ref{subsec:detail} for quantitative analysis).

Let $N_{mem}$ denote the number of memory frames. We first utilize the tracker to predict $N_{mem}$ intermediate bounding boxes in the memory frames for the template $z_{t}$. 
We adopt PrPool~\cite{DBLP:conf/eccv/JiangLMXJ18} to pool $N_{mem}$ features based on the intermediate boxes. Then we use the pooled features as templates to conduct multi-scale correlation~\cite{DBLP:conf/eccv/ZhangPFLH20} with the deep feature of $x_{t}$. Note that the original classification branch and the memory branch share the same weights in terms of the multi-scale correlation module. All $N_{mem}$ correlation maps, denoted as $\left\{ C^{u}_{corr} \mid 1 \le u \le N_{mem} \right\}$, are integrated together by a confidence-value strategy. Specifically, given a correlation map $C^{u}_{corr}$, we utilize two $3 \times 3$ convolution layers to generate a confidence map $C^{u}_{conf}$ and a value map $C^{u}_{val}$ with the same dimension. 
We then normalize $C^{u}_{conf}$ element-wise across all confidence maps as weights on $C^{u}_{val}$. The finally integrated correlation map $C$ is formulated as follows:
\begin{equation}\label{eqn:integration}
C = \sum_{1 \le u \le N_{mem}}{\operatorname{softmax}{(C_{conf}^{u})} \odot C_{val}^{u}},
\end{equation}
where $\odot$ denotes Hadamard product.
As shown in Fig.~\ref{fig:network}, the integrated map $C$ is converted to $25 \times 25 \times 1$ via convolution, yielding the response map $\mathcal{R}_{mem}$ of the object in search area $x_{t}$. The total loss function $\mathcal{L}$ for training is:  
\begin{equation}\label{eqn:loss_stage3}
\mathcal{L} = \mathcal{L}_{reg} + \lambda_{1}\mathcal{L}_{cls} + \lambda_{2}\mathcal{L}_{mem},
\end{equation}
where $\lambda_{1}$ and $\lambda_{2}$ are weight parameters. We use the BCE loss~\cite{DBLP:journals/anor/BoerKMR05} as the cycle memory loss $\mathcal{L}_{mem}$, which shares the same pseudo ground truth label as in $\mathcal{L}_{cls}$.

\begin{table*}[ht]
	\small
	\begin{center}
		\caption{Results on the VOT benchmarks. The proposed trackers perform well over the state-of-the-art unsupervised trackers. Boldface denotes the best performance among all trackers built without video labels for offline training (the same below). } \label{tab:vot}	
		\vspace{-0.1in}
		\begin{tabular*} {1.0\textwidth} {@{\extracolsep{\fill}}lcccccccccc}
			\toprule
			\multirow{2}*{\ \ Tracker} & \multirow{2}*{Unsup.} & 
			\multicolumn{3}{c}{VOT2016} 
			& \multicolumn{3}{c}{VOT2017/18}  & \multicolumn{3}{c}{VOT2020}  \\
			\cmidrule(r){3-5}  \cmidrule(r){6-8} \cmidrule(r){9-11}
			 &  & A $\uparrow$ & R $\downarrow$ & EAO $\uparrow$ & A $\uparrow$ & R $\downarrow$ & EAO $\uparrow$ & A $\uparrow$ & R $\uparrow$ & EAO $\uparrow$ \\
			\midrule
			\ \ SiamFC~\cite{DBLP:conf/eccv/BertinettoVHVT16} & No & 0.532 & 0.461 & 0.235 & 0.503 & 0.585 & 0.188 & 0.418 & 0.502 & 0.179 \\
			\ \ DaSiamRPN~\cite{DBLP:conf/eccv/ZhuWLWYH18} & No & 0.61 & 0.22 & 0.411 & 0.56 & 0.34 & 0.326 & - & - & - \\
			\ \ ATOM~\cite{DBLP:conf/cvpr/DanelljanBKF19} & No & - & - & - & 0.590 & 0.204 & 0.401 & 0.462 & 0.734 & 0.271 \\
			\ \ DiMP~\cite{DBLP:conf/iccv/BhatDGT19} & No & - & - & - & 0.597 & 0.153 & 0.440 & 0.457 & 0.740 & 0.274 \\
			\midrule
			\ \ IVT~\cite{DBLP:journals/ijcv/RossLLY08} & Yes & 0.419 & 1.109 & 0.115 & 0.400 & 1.639 & 0.076 & 0.345 & 0.244 & 0.092 \\
			\ \ MIL~\cite{DBLP:journals/pami/BabenkoYB11} & Yes & 0.407 & 0.727 & 0.165 & 0.394 & 1.011 & 0.118 & 0.367 & 0.322 & 0.113 \\
			\ \ KCF~\cite{DBLP:journals/pami/HenriquesC0B15} & Yes & 0.489 & 0.569 & 0.192 & 0.447 & 0.773 & 0.135 & 0.407 & 0.432 & 0.154 \\
			\ \ ECO~\cite{DBLP:conf/cvpr/DanelljanBKF17} & Yes & 0.55 & \textbf{\textcolor{black}{0.20}} & 0.375 & 0.48 & \textbf{\textcolor{black}{0.27}} & 0.280 & - & - & - \\ 
			\midrule
			\ \ S$^2$SiamFC~\cite{DBLP:conf/mm/SioMSCC20} & Yes & 0.493 & 0.639 & 0.215 & 0.463 & 0.782 & 0.180 & - & - & - \\
			\ \ LUDT~\cite{DBLP:journals/ijcv/WangZSMLL21} & Yes & 0.544 & 0.422 & 0.232 & 0.463 & 0.693& 0.154 & - & - & - \\
			\ \ LUDT+~\cite{DBLP:journals/ijcv/WangZSMLL21} & Yes & 0.570 & 0.331 & 0.299 & 0.490 & 0.412 & 0.230 & - & - & - \\
			\midrule
			\ \  USOT \ (Ours) & Yes & 0.593 & 0.336 & 0.351 & 0.564 & 0.435 & 0.290 & \textbf{\textcolor{black}{0.458}} & \textbf{\textcolor{black}{0.600}} & \textbf{\textcolor{black}{0.222}} \\
			\ \  USOT* (Ours) & Yes & \textbf{\textcolor{black}{0.600}} & 0.233 & \textbf{\textcolor{black}{0.402}} & \textbf{\textcolor{black}{0.578}} & 0.304 & \textbf{\textcolor{black}{0.344}} & 0.448 & \textbf{\textcolor{black}{0.600}} & 0.219 \\
			\bottomrule
		\end{tabular*}
	\end{center}
	\vspace{-0.2in}
\end{table*}

\section{Experiments}

This section presents the results of our unsupervised tracker on multiple benchmarks, with comparisons to the state-of-the-art tracking algorithms. Extensive ablation studies are provided to analyze the effectiveness of our design choices.

\subsection{Implementation Details}

\noindent\textbf{Training.} Our tracker is trained on the data collected from the training sets of four datasets including GOT-10k~\cite{huang2019got}, ImageNet VID~\cite{DBLP:journals/ijcv/RussakovskyDSKS15}, LaSOT~\cite{DBLP:conf/cvpr/FanLYCDYBXLL19} and YouTube-VOS~\cite{DBLP:conf/eccv/XuYFYYLPCH18}. Note that the ground-truth labels of these training sets are not available in our method. The hyper-parameters for extracting the video box sequences are set as $\alpha=0.3$, $\beta=0.5$, $\gamma=4.1$ respectively. Our network adopts ResNet-50~\cite{DBLP:conf/cvpr/HeZRS16} as the backbone network, and uses the third convolutional block to extract deep features for input images. We notice that existing CNN backbones pretrained on the ImageNet dataset~\cite{DBLP:journals/ijcv/RussakovskyDSKS15} contain information from manual labels. 
For the sake of solid comparisons, we conduct all the experiments in two settings (i.e., w/o and w/ supervised backbone pretraining on ImageNet). During training, we use synchronized SGD~\cite{DBLP:journals/neco/LeCunBDHHHJ89} on $4$ NVIDIA GeForce RTX 3090 GPUs. Each GPU hosts $12$ groups of training instances. The whole end-to-end training phase takes $30$ epochs in total, in which cycle memory is conducted only within the last $25$ epochs. We start with a warm-up learning rate from $2.5 \times 10^{-3}$ to $5 \times 10^{-3}$ in the first $6$ epochs, while the remaining epochs adopt an exponentially decreasing learning rate from $5 \times 10^{-3}$ down to $2 \times 10^{-5}$. 

At each training step, we sample a template bounding box from $\mathcal{B}^{\prime}$ with $\theta_{1} = 0.4$ and crop a template patch and a search area as SiamFC~\cite{DBLP:conf/eccv/BertinettoVHVT16}. This image pair is augmented with horizontal and vertical flips. For training the online module with cycle memory, we input extra $N_{mem}=4$ memory search areas together with the template-search pair. Other hyper-parameters for cycle memory are set to $\gamma=2.5$, $\theta_{2} = 0.45$ and $\theta_{3} = 0.40$. The trade-off parameter in the loss function is fixed to $\lambda_{1}=0.2$ for naive Siamese training, and for cycle memory we keep $\lambda_{1}+\lambda_{2}=0.9$ and 
gradually decrease $\lambda_{1}$ from $0.3$, in order to make the tracker gently suit to the tracking task in longer temporal spans. 

\noindent\textbf{Inference.} The inference is performed on both the offline and online branches. The offline branch follows the conventional inference methodology of Siamese networks \cite{DBLP:conf/cvpr/LiWWZXY19,DBLP:conf/eccv/ZhangPFLH20}. Based on the template feature from frame $I_{1}$, the offline response maps $\mathcal{R}_{cls}$ and bounding boxes $\mathcal{R}_{reg}$ in subsequent frames are generated by the offline classifier and regressor. 
On the other hand, the online branch maintains a memory queue of templates and dynamically updates this queue with new predictions. In practice, when tracking the object in frame $I_{t}$, the memory queue consists of totally $N_q$ templates, including two ground-truth templates from frame $I_1$ (i.e., the original template and its horizontal flip), the latest predicted template from frame $I_{t-1}$, and $N_{q}-3$ historical templates with the highest scores in $\mathcal{R}$ from frame $I_{2}$ to $I_{t-2}$. The final response map $\mathcal{R}$ is a weighted sum of $\mathcal{R}_{cls}$ and $\mathcal{R}_{mem}$, namely $\mathcal{R} = (1-w)\mathcal{R}_{cls} + w \mathcal{R}_{mem}$, where the weight $w$ is set to $0.7$ throughout our experiments.

\subsection{State-of-the-art Comparison}

We compare our method with the state-of-the-art unsupervised and supervised trackers. The comparisons are conducted on six benchmarks, including VOT2016~\cite{DBLP:conf/eccv/KristanLMFPCVHL16}, VOT2017/18~\cite{DBLP:conf/eccv/KristanLMFPZVBL18}, VOT2020~\cite{DBLP:conf/eccv/KristanLMFPKDZL20}, TrackingNet~\cite{DBLP:conf/eccv/MullerBGAG18}, OTB2015~\cite{DBLP:journals/pami/WuLY15} and LaSOT~\cite{DBLP:conf/cvpr/FanLYCDYBXLL19}. 
We denote by USOT the proposed tracker trained with the unsupervised backbone initialization~\cite{MocoV2}, while denoting by USOT* that with supervised ImageNet pretraining. It is worth mentioning that we keep all the hyper-parameters fixed throughout the experiments to report results on all the datasets.

\noindent\textbf{VOT2016.} The VOT2016 dataset contains $60$ video sequences. We adopt the Accuracy (A), Robustness (R) and Expected Average Overlap (EAO)~\cite{DBLP:conf/eccv/KristanLMFPCVHL16} as the VOT-toolkit to evaluate the overall performance. Tab.~\ref{tab:vot} shows that the proposed trackers, both USOT and USOT*, significantly outperform the state-of-the-art unsupervised trackers, with respectively $5.2$ and $10.3$ points increase in EAO over the top-performing unsupervised tracker LUDT+.

\noindent\textbf{VOT2017/18.} The VOT2017/18 dataset consists of $60$ more challenging video sequences. Tab.~\ref{tab:vot} shows that our USOT* obtains $8.8$ and $11.4$ points increase respectively in Accuracy and EAO compared with LUDT+. Our USOT without supervised backbone still outperforms LUDT+ by respectively $7.4$ and $6.0$ points in Accuracy and EAO.

\noindent\textbf{VOT2020.} The VOT2020 dataset encodes targets with segmentation masks, and updates the calculation of Accuracy (A), Robustness (R) and Expected Average Overlap (EAO)~\cite{DBLP:conf/eccv/KristanLMFPKDZL20}. Tab.~\ref{tab:vot} shows that our USOT* outperforms SiamFC by $3.0$, $9.8$ and $4.0$ points respectively in A, R and EAO. Our USOT even achieves better performance compared with USOT* with an EAO of $0.222$. 

\begin{table}
	\small
	\begin{center}
		\caption{Results on the TrackingNet~\cite{DBLP:conf/eccv/MullerBGAG18} dataset. The proposed unsupervised trackers, USOT and USOT*, perform well over the state-of-the-arts.} \label{tab:trackingnet}	
		\vspace{-0.1in}
		\begin{tabular*} {0.48\textwidth} {@{\extracolsep{\fill}}lcccc}
			\toprule
			\ \ Trackers & Unsup. & Succ. $\uparrow$  & Prec. $\uparrow$
			& Norm P. $\uparrow$\\
			\midrule
			\ \ SiamFC~\cite{DBLP:conf/eccv/BertinettoVHVT16} & No & 57.1 & 53.3 & 66.3  \\
			\ \ MDNet~\cite{DBLP:conf/cvpr/NamH16} & No & 61.4 & 55.5 & 71.0  \\
			\ \ UpdateNet~\cite{DBLP:conf/iccv/ZhangGWDK19} & No & 67.7 & 62.5 & 75.2  \\
			\ \ ATOM~\cite{DBLP:conf/cvpr/DanelljanBKF19} & No & 70.3 & 64.8 & 77.1 \\
			\ \ SiamRPN++~\cite{DBLP:conf/cvpr/LiWWZXY19} & No & 73.3 & 69.4 & 80.0 \\ 
			\midrule
			\ \ KCF~\cite{DBLP:journals/pami/HenriquesC0B15} & Yes & 44.7 & 41.9 & 54.6 \\
			\ \ DSST~\cite{DBLP:conf/bmvc/DanelljanHKF14} & Yes & 46.4 & 46.0 & 58.8 \\	
			\ \ ECO~\cite{DBLP:conf/cvpr/DanelljanBKF17} & Yes & 56.1 & 48.9 & 62.1  \\
			\midrule
			\ \ LUDT~\cite{DBLP:journals/ijcv/WangZSMLL21} & Yes & 54.3 & 46.9 & 59.3 \\
			\ \ LUDT+~\cite{DBLP:journals/ijcv/WangZSMLL21} & Yes & 56.3 & 49.5 & 63.3 \\
			\midrule
			\ \  USOT \ (Ours) & Yes & 59.9 & 55.1 & 68.2 \\
			\ \  USOT* (Ours) & Yes & \textbf{\textcolor{black}{61.5}} & \textbf{\textcolor{black}{56.6}} & \textbf{\textcolor{black}{69.1}} \\
			\bottomrule
		\end{tabular*}
	\end{center}
	\vspace{-0.25in}
\end{table}

\noindent\textbf{TrackingNet.} The TrackingNet dataset contains over $30000$ videos, with $511$ videos for testing. Tab.~\ref{tab:trackingnet} shows that our USOT* outperforms LUDT+ with $5.2$ points in Success and $7.1$ points in Precision. Our USOT is comparable with USOT* on the TrackingNet dataset. This suggests that unsupervised representation learning has the potential to perform as well as supervised ImageNet pretraining. 

\noindent\textbf{OTB2015.} The OTB2015 dataset contains $100$ videos. Tab.~\ref{tab:OTB2015} shows that our USOT outperforms USOT*. This affirms the potential of unsupervised representation learning from scratch. Furthermore, the proposed unsupervised trackers achieve comparable performance with LUDT and SiamFC. Our trackers perform slightly worse than LUDT+, because LUDT+ adopts some online tracking techniques in ~\cite{DBLP:conf/cvpr/DanelljanBKF17} over LUDT for better performance. 

\begin{table*}
	\small
	\begin{center}
		\caption{Results on the OTB2015~\cite{DBLP:journals/pami/WuLY15} dataset. Our unsupervised trackers achieve comparable results with the previous supervised and unsupervised trackers.}  \label{tab:OTB2015}
		\vspace{-0.1in}
		\begin{tabular*} {1.0\textwidth} {@{\extracolsep{\fill}}lccccc|cc|cc|cc}
			\toprule
			\ \ \multirow{2}*{Tracker}  & DCFNet & SiamFC & SiamRPN & ATOM & DiMP & DSST & KCF & LUDT & LUDT+ & USOT & USOT* \\
			& \cite{DBLP:journals/corr/WangGXZH17} &
			\cite{DBLP:conf/eccv/BertinettoVHVT16} &
			\cite{DBLP:conf/cvpr/LiYWZH18} &
			\cite{DBLP:conf/cvpr/DanelljanBKF19} & 
			\cite{DBLP:conf/iccv/BhatDGT19} & 
			\cite{DBLP:conf/bmvc/DanelljanHKF14} & 
			\cite{DBLP:journals/pami/HenriquesC0B15} & 
			\cite{DBLP:journals/ijcv/WangZSMLL21} &
			\cite{DBLP:journals/ijcv/WangZSMLL21} & (Ours) & (Ours) \\
			\midrule
			\ \ AUC success & 58.0 & 58.2 & 63.7 & 66.7 & 68.6 & 51.8 & 48.5  & 60.2 &  \textbf{63.9} & 58.9 & 57.4 \\
			\ \ Distance precision  & - & 77.1 & 85.1 & 87.9 & 89.9 & 68.9 & 69.6  & 76.9 & \textbf{84.3} & 80.6 & 77.5 \\
			\bottomrule
		\end{tabular*}
	\end{center}
	\vspace{-0.25in}
\end{table*}

\begin{figure}
	\begin{center}
		\includegraphics[width=0.50\textwidth]{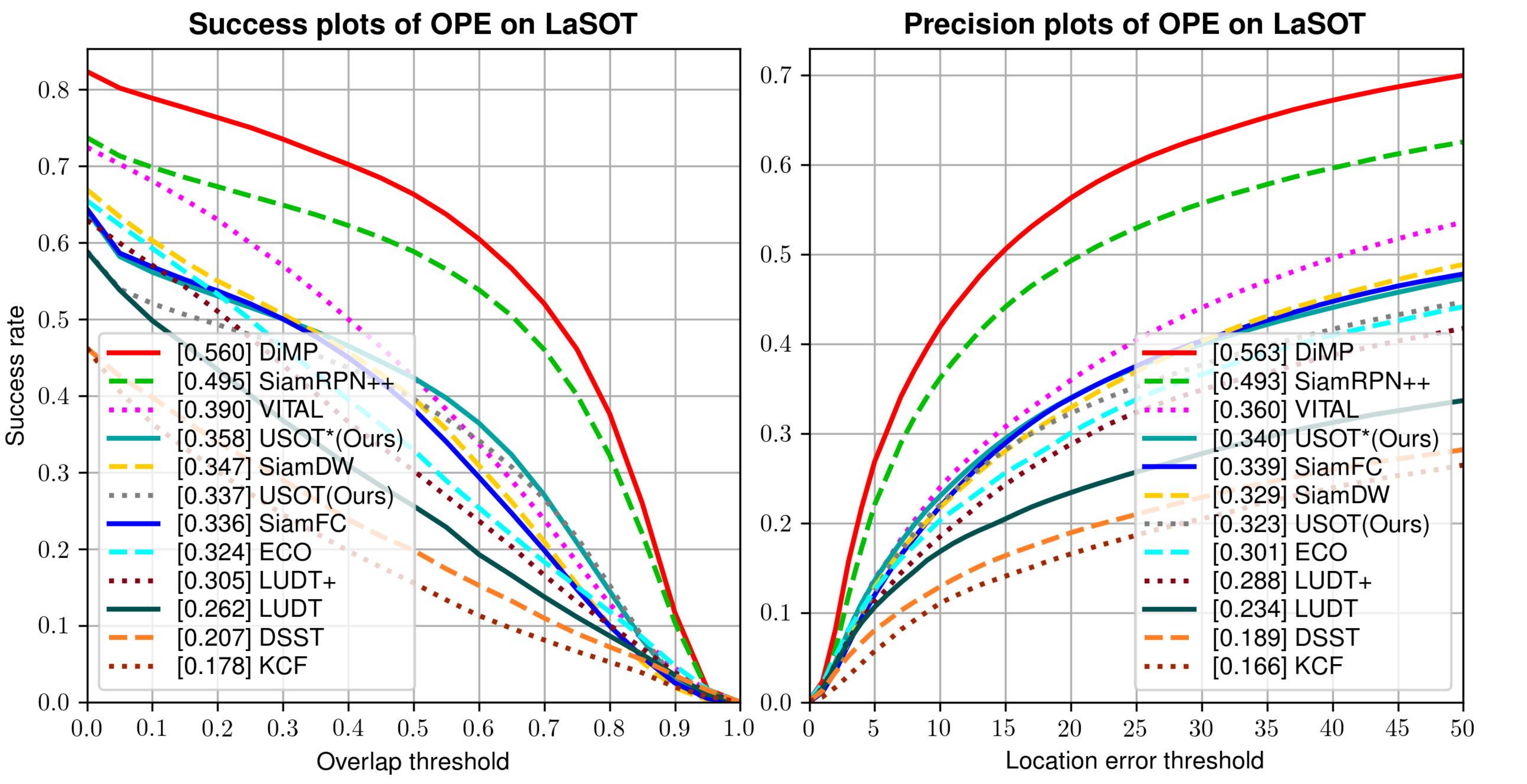}
	\end{center}
	\vspace{-0.25in}
	\caption{Success plot and Precision plot on the LaSOT testing set~\cite{DBLP:conf/cvpr/FanLYCDYBXLL19}.} 
	\vspace{-0.15in}
	\label{fig:lasot}
\end{figure}

\noindent\textbf{LaSOT.} The LaSOT testing set consists of $280$ videos with an average length over $2000$. LaSOT is important for measuring long-term tracking performance. Fig.~\ref{fig:lasot} shows that our USOT* significantly outperforms LUDT+ with an increase of $5.3$ and $5.2$ on Success and Precision, respectively. Our USOT achieves comparable results to the supervised trackers SiamFC~\cite{DBLP:conf/eccv/BertinettoVHVT16} and SiamDW~\cite{DBLP:conf/cvpr/ZhangP19}.

\subsection{Ablation Studies}\label{subsec:detail}

\noindent\textbf{Training Stage Indispensability.} 
We do extensive ablation studies on the importance of different stages in the training phase. Experiments are conducted on the VOT2017/18 benchmark with USOT*. Tab.~\ref{tab:indispensibility} shows that training a naive tracker from single-frame pairs with random cropping causes a significant accuracy drop compared with our proposed box sequence generation. In addition, directly conducting cycle memory training without the naive Siamese tracker initialization also causes a large performance drop. 

\begin{table}[t]
	\small
	\begin{center}
		\caption{ Ablation studies on our pseudo box generation module and naive Siamese tracker initialization before cycle memory training on the VOT2017/18 dataset.} \label{tab:indispensibility}
		\vspace{-0.1in}
		\begin{tabular*}{0.43\textwidth} {@{\extracolsep{\fill}}cccc}
			\toprule
			Operations & {A $\uparrow$}& {R $\downarrow$} & {EAO $\uparrow$}\\
			\midrule
			Random boxes & 0.488 & 0.646 & 0.195 \\
			Our gnereated boxes  & \textbf{0.567} & \textbf{0.520} &  \textbf{0.263}\\
			\midrule
			\ \ w/o naive Siamese learning& 0.575 & 0.389 &  0.306 \\
			\ \ w/\ \ \ naive Siamese learning& \textbf{0.578} & \textbf{0.304} & \textbf{0.344} \\
			\midrule
		\end{tabular*}
	\end{center}
	\vspace{-0.16in}
\end{table}

\noindent\textbf{Video Utilization Rate.} 
We use the video quality score $Q_{v}(\mathcal{I})$ in Eqn.~\ref{eqn:video_score} to filter out noisy sequences during unsupervised training. On the GOT-10k dataset and the VID dataset, we utilize 50.8\% and 56.3\% videos respectively within all videos available for training, covering rich varieties of unlabeled videos. 

\noindent\textbf{Frame Interval.} 
Our proposed training method can learn the appearance information across long intervals. This facilitates unsupervised trackers to adapt to temporal appearance changes. Compared to the very short frame intervals in previous deep unsupervised trackers S$^2$SiamFC~\cite{DBLP:conf/mm/SioMSCC20} (i.e., $0$ frame) and UDT~\cite{DBLP:conf/cvpr/WangS0ZLL19} (i.e., $<10$ frames), the training instances sampled by our method possess an averaged long frame interval of $41.1$ and $64.6$ respectively on the GOT-10k and VID datasets.

\begin{table}
	\small
	\begin{center}
		\caption{Quantitative results on the IoU success rates of the pseudo boxes in template frames and memory frames.} \label{tab:quant}	
		\vspace{-0.1in}
		\begin{tabular*} {0.43\textwidth} {@{\extracolsep{\fill}}lcccccc}
			\toprule
			\ \ \multirow{2}{*}{Dataset $\backslash$ IoU} & \multicolumn{2}{c}{Template} & \multicolumn{2}{c}{Memory}  \\
			\cmidrule{2-3} \cmidrule{4-5}
			 & 0.3 & 0.5 & 0.3 & 0.5 \\
			\midrule
			\ \ GOT-10k & 63.2\% & 45.5\% & 62.0\% & 43.8\% \\
			\ \ VID  & 64.4\% & 42.1\% & 63.9\% & 42.0\% \\
			\bottomrule
		\end{tabular*}
	\end{center}
	\vspace{-0.2in}
\end{table}

\noindent\textbf{Pseudo Bounding Box Generation.} To better investigate the precision of the pseudo bounding boxes, we collect over $10^{4}$ training instances and compute the IoU scores between the output pseudo bounding boxes and the ground truth bounding boxes on both the GOT-10k and VID datasets. Tab.~\ref{tab:quant} shows the success rates of the pseudo bounding boxes over different IoU scores in both template frames and memory frames. 
On both datasets, over $63\%$ sampled boxes in template frames cover at least parts of the foreground objects ($IoU > 0.3$), while over $42\%$ sampled boxes in template frames are precise enough to cover approximately the entire objects ($IoU > 0.5$). Besides, from the small difference between the IoU success rates of pseudo boxes in template frames and memory frames on both datasets, we conclude that using large frame intervals for cycle memory training only slightly decreases the reliability of memory frames compared to template frames. This explains why our unsupervised tracker can learn from large motions.

\noindent\textbf{Training Dataset.} Since most existing unsupervised deep trackers are trained on the VID dataset, we investigate the impact of training data on USOT* on the VOT2017/18 benchmark. 
As is shown in Tab.~\ref{tab:dataset}, when only using VID as the training set, the proposed tracker still achieves $0.315$ in EAO, with an $8.5$ points increase over the state-of-the-art unsupervised tracker LUDT+ (i.e., $0.230$ in EAO). Besides, our tracker benefits from training on more unlabeled videos, inferring the great potential of unsupervised tracking.

\begin{table}[t]
	\small
	 \setlength\tabcolsep{1pt}
	\begin{center}
		\caption{Ablation studies on training data. With more unlabeled videos used for training, the proposed USOT* achieves better results on the VOT2017/18 dataset. } \label{tab:dataset}
		\vspace{-0.1in}
		\begin{tabular*}{0.45\textwidth} {@{\extracolsep{\fill}}ccccccc}
			\toprule
			\multicolumn{4}{c}{Training Data}& \multirow{2}*{A $\uparrow$} & \multirow{2}*{R $\downarrow$}  & \multirow{2}*{EAO $\uparrow$} \\
			\cmidrule(r){1-4} 
			\ \ VID & GOT-10k & LaSOT & YT-VOS & ~ & ~ & ~ \\
			\midrule
			\ \ \CheckmarkBold & & & & 0.576 & 0.337 & 0.315 \\
			\ \ \CheckmarkBold & \CheckmarkBold & & & \textbf{0.587} & 0.323 & 0.320 \\
			\ \ \CheckmarkBold & \CheckmarkBold & \CheckmarkBold & & 0.579 & 0.328 & 0.337 \\
			\ \ \CheckmarkBold & \CheckmarkBold & \CheckmarkBold & \CheckmarkBold &  0.578 & \textbf{0.304} & \textbf{0.344} \\
			\midrule
		\end{tabular*}
	\end{center}
	\vspace{-0.25in}
\end{table}

\begin{table}[t]
	\small
	\begin{center}
		\caption{Parameter sensitivity of the length and weight of the online memory queue on the VOT2017/18 dataset.} \label{tab:memory}
		\vspace{-0.1in}
		\begin{tabular*} {0.45\textwidth} {@{\extracolsep{\fill}}cccccc}
			\toprule
			\ $N_{q}$ $\backslash$ $w$ & 0.3 & 0.5 & 0.6 & 0.7 & 0.8  \\
			\midrule
			\ 5  & 0.289 & 0.302 & 0.323 & 0.313 & 0.323  \\
			\ 6  & 0.294 & 0.312 & 0.312 & 0.329 & 0.322 \\
			\ 7  & 0.310 & 0.318 & 0.336 & \textbf{0.344} & 0.331 \\
			\ 8  & 0.302 & 0.300 & 0.319 & 0.341 & 0.338 \\
			\bottomrule
		\end{tabular*}
	\end{center}
	\vspace{-0.25in}
\end{table}

\noindent\textbf{Online Update.} We study the parameter sensitivity of $N_{q}$ and $w$ in the online memory module. $N_{q}$ indicates the number of memorized features collected online in the memory queue, while $w$ indicates the weight for $\mathcal{R}_{mem}$. Tab.~\ref{tab:memory} reports the EAO scores of USOT* on the VOT2017/18 dataset. The cooperation of offline and online modules with $w=0.7$ benefits the proposed tracker most, and setting the length of the memory queue $N_{q}$ to $7$ is most suitable.

\section{Concluding Remarks}

In this paper, we propose learning a robust tracker from unlabeled videos from scratch. 
We first generate candidate box sequences to cover moving objects in videos. We then train a naive Siamese tracker using single-frame pairs. We finally continue training the naive tracker in longer temporal spans with a novel cycle memory scheme, enabling the tracker to update online. Extensive experiments demonstrate that the proposed unsupervised tracker sets new state-of-the-art unsupervised tracking results, and even performs on par with recent supervised deep trackers. This work unveils the power of unsupervised learning for object tracking. 

\vspace{1mm}
\noindent\textbf{Acknowledgements.} This work was supported by NSFC (61906119, U19B2035), Shanghai Municipal Science and Technology Major Project (2021SHZDZX0102), and Shanghai Pujiang Program. 

{\small
\bibliographystyle{bib/ieee_fullname}
\bibliography{bib/egbib}
}

\end{document}